\documentclass{article} 
\usepackage{iclr2022_conference,times}

\usepackage{amsmath,amsfonts,bm}

\def\eqref#1{equation~\ref{#1}}

\def\1{\bm{1}}

\def\vx{{\bm{x}}}

\DeclareMathAlphabet{\mathsfit}{\encodingdefault}{\sfdefault}{m}{sl}
\SetMathAlphabet{\mathsfit}{bold}{\encodingdefault}{\sfdefault}{bx}{n}

\def\gI{{\mathcal{I}}}

\def\gL{{\mathcal{L}}}

\def\gT{{\mathcal{T}}}

\def \xi{\vx^I}
\def \xt{\vx^T}

\newcommand{\R}{\mathbb{R}}

\newtheorem{remark}{\textbf{Remark}}

\usepackage[colorlinks=true]{hyperref}
\usepackage{url}
\usepackage{amsfonts}
\usepackage{amssymb}
\usepackage{multirow}
\usepackage{graphicx}
\usepackage{makecell}
\usepackage{booktabs}
\title{FILIP: Fine-grained Interactive Language-Image Pre-Training}

\author{Lewei Yao $^{1,2}$\thanks{Equal contribution} 
\hspace{3mm}Runhui Huang$^{3*}$ 
\hspace{3mm}Lu Hou$^{1*}$ 
\hspace{3mm}Guansong Lu$^{1}$ 
\hspace{3mm}Minzhe Niu$^{1}$ \\
\textbf{
\hspace{-1mm}Hang Xu$^{1}$\thanks{Corresponding authors: xu.hang@huawei.com, xdliang328@gmail.com}
\hspace{3mm}Xiaodan Liang $^{3\dag}$ 
\hspace{3mm}Zhenguo Li$^{1}$ 
\hspace{3mm}Xin Jiang$^{1}$ 
\hspace{3mm}Chunjing Xu$^{1}$} \\ 
\hspace{0.03in} $^{1}$Huawei Noah’s Ark Lab \\ 
\hspace{0.03in} $^{2}$Hong Kong University of Science and Technology \\ 
\hspace{0.03in} $^{3}$Sun Yat-sen University\\
}

\iclrfinalcopy 
\begin{document}

\maketitle

\begin{abstract}

Unsupervised large-scale vision-language pre-training has shown promising advances on various downstream tasks. Existing methods often  model the cross-modal interaction either via the similarity of the global feature of each modality which misses sufficient information, or finer-grained interactions using cross/self-attention upon visual  and textual tokens. However, cross/self-attention suffers from inferior efficiency in both training and inference. In this paper, we introduce a large-scale Fine-grained Interactive Language-Image Pre-training (FILIP) to achieve finer-level alignment through a cross-modal late interaction mechanism, which uses a token-wise maximum similarity between visual and textual tokens to guide the contrastive objective. FILIP successfully leverages the finer-grained expressiveness between image patches and textual words by modifying only contrastive loss, while simultaneously gaining the ability to pre-compute image and text representations offline at inference, keeping both large-scale training and inference efficient. Furthermore, we construct a new large-scale image-text pair dataset called FILIP300M for pre-training. Experiments show that FILIP achieves state-of-the-art performance on multiple downstream vision-language tasks including zero-shot image classification and image-text retrieval. The visualization on word-patch alignment further shows that FILIP can learn meaningful fine-grained features with promising localization ability. 

\end{abstract}

\section{Introduction}
Large-scale Vision-Language Pre-training (VLP) models like CLIP \citep{radford2021learning} and ALIGN  \citep{jia2021scaling} have recently demonstrated success across 
various downstream tasks. They learn visual and textual representations from millions of image-text  pairs collected from the Internet and show superior zero-shot ability and robustness. The core technique of these models lies in the global contrastive alignment of the images and texts through a dual-stream model. Such architecture is inference-efficient for downstream tasks like retrieval because the encoders for the two modalities can be decoupled and the image or text representations can be pre-computed offline. 
However, CLIP and ALIGN
model the  cross-modal interaction via solely the similarity of the global feature of each modality, lacking the ability of capturing finer-level information like the relationship between visual objects and textual words.
In this paper, we develop a simple yet efficient cross-modal finer-grained interaction mechanism for large-scale VLP.

To achieve finer-grained cross-modal interaction, previous methods mainly exploited two kinds of methods.
(1) One line of work \citep{chen2020uniter,li2020oscar,m5product,li2020unimo,zhang2021vinvl,capture} uses a pre-trained object detector to extract region-of-interest (ROI) features from images, and then fuses it with the paired text through a VLP model.
This design complicates the pre-training due to  pre-computing and storing  a large number of ROI features. 
In addition, the zero-shot ability of these approaches is usually limited by the predefined number of classes and their performance is also restricted by the quality of the detector.
(2) Another line of work \citep{li2021align, kim2021vilt} 
enforces the token-wise or patch-wise representations from both modalities into the same space and models these finer-grained interactions via cross-attention \citep{li2021align} or self-attention \citep{kim2021vilt}. 
However, these methods are usually less efficient in terms of both training and inference. In particular, during training, cross-attention in \citep{li2021align} requires to be performed in an encoder-decoder structure, while the complexity of the self-attention \citep{kim2021vilt} grows quadratically with the length of the prolonged concatenated sequences of both modalities. During inference, the data from both modalities are intertwined to compute the cross-attention or self-attention, and can not be pre-computed offline as dual-stream models like CLIP and ALIGN.
This can be less efficient for downstream tasks like image/text retrieval and image classification.

In this paper, we propose a large-scale Fine-grained Interactive Language-Image Pre-training framework named FILIP 
to address these limitations.
Inspired by \cite{khattab2020colbert}, 
we model  the fine-grained semantic alignment  through a novel cross-modal
late interaction mechanism in the contrastive loss, instead of using cross or self-attention.
Specifically, our fine-grained contrastive learning uses a token-wise maximum similarity between visual  and textual tokens to guide the contrastive objective.
In this way,
FILIP  successfully leverages the finer-grained expressiveness
among image patches and textual words 
while simultaneously gaining the ability to pre-compute image and text representations offline.
Unlike \cite{khattab2020colbert}, we discard the padded tokens and use average 
instead summation 
of token-wise maximum similarities when computing the image-text alignment,
which enhances the cross-modal representation learning and stabilizes training.
Furthermore, we construct a large-scale pre-training dataset named FILIP300M from the Internet.
Data cleaning and image-text data augmentation are also explored  and proved useful in this work.

Extensive experiments show that by effectively learning fine-grained representations,
FILIP achieves state-of-the-art performance on multiple downstream vision-language tasks, including zero-shot image classification and image-text retrieval. For example, FILIP reaches  77.1\% top-1 accuracy for zero-shot ImageNet classification, surpassing  CLIP with less training data.
Visualizations on  word-patch alignment further show that FILIP learns meaningful finer-grained features with promising localization ability. 



\section{Related Work}
\paragraph{Vision-Language Pre-training Models.} The pre-train-and-fine-tune scheme has achieved great success in the domains of natural language processing~\citep{devlin2018bert,brown2020language} and computer vision~\citep{dosovitskiy2020image}.
It is then naturally extended to a joint cross-modal domain of 
Vision-and-Language Pre-training (VLP). 
The pre-training datasets of recent VLP models 
include publically available 
datasets 
like YFCC100M \citep{thomee2016yfcc100m} and CC12M \citep{changpinyo2021cc12m}, as well as 
larger-scale datasets with more than 100M samples
in CLIP \citep{radford2021learning} and ALIGN \citep{jia2021scaling}, which are shown to be even more powerful. 
The pre-training tasks of VLP models can be categorized into two categories: image-text contrastive learning task and Language Modeling (LM) based tasks:
(i) CLIP \citep{radford2021learning}, ALIGN \citep{jia2021scaling} and UNIMO \citep{li2020unimo} make use of cross-modal contrastive learning which aligns the textual and visual information into a unified semantic space; (ii) VisualBERT \citep{li2019visualbert}, UNITER \citep{chen2020uniter}, M6 \citep{lin2021m6}, and DALL-E \citep{ramesh2021zeroshot} employ LM-like objectives, including both masked LM (e.g., Masked Language/Region Modeling), and autoregressive LM (e.g., image captioning, text-grounded image generation).
On the other hand, some methods rely on a pre-trained object detection model such as Faster-RCNN \citep{ren2015faster} to extract image regional
features offline, which requires extra labeled bounding-box data and makes the approach less scalable.
Recent efforts such as SOHO \citep{huang2021seeing} and SimVLM \citep{wang2021simvlm} try to eliminate this burden via visual dictionary or PrefixLM 
\citep{raffel2020exploring}. 
In this paper, we 
 directly 
learn fine-grained vision-language representations
in an end-to-end and simpler manner while maintaining the benefit of inference efficiency.

\paragraph{Multi-Modality Interaction Mechanism.} 
The core of vision-language pre-training models lies in modeling the interaction between the two modalities. 
There are mainly two types of cross-modal interaction architectures: single-stream and dual-stream models. 
Single-stream models like VisualBERT \citep{li2019visualbert} and ViLT \citep{kim2021vilt} directly concatenate the patch-wise or regional visual features and textual embeddings 
and feed them to the transformer-based model.
Dual-stream models such as ViLBERT \citep{lu2019vilbert} and CLIP~\citep{radford2021learning} have separate encoders for different modalities. 
This allows flexible use of different models for different modalities, and
 efficient inference for downstream tasks like image-text retrieval, through the ability of decoupling the encoders and pre-compute image/text features offline.
In this paper, while following the dual-stream approach for its flexible and efficient inference,
we further propose a new multi-modal interaction mechanism to capture the fine-grained representations. 


\vspace{-1mm}
\section{Method}

\begin{figure}
\vspace{-3mm}
\begin{center}
\includegraphics[width=\textwidth]{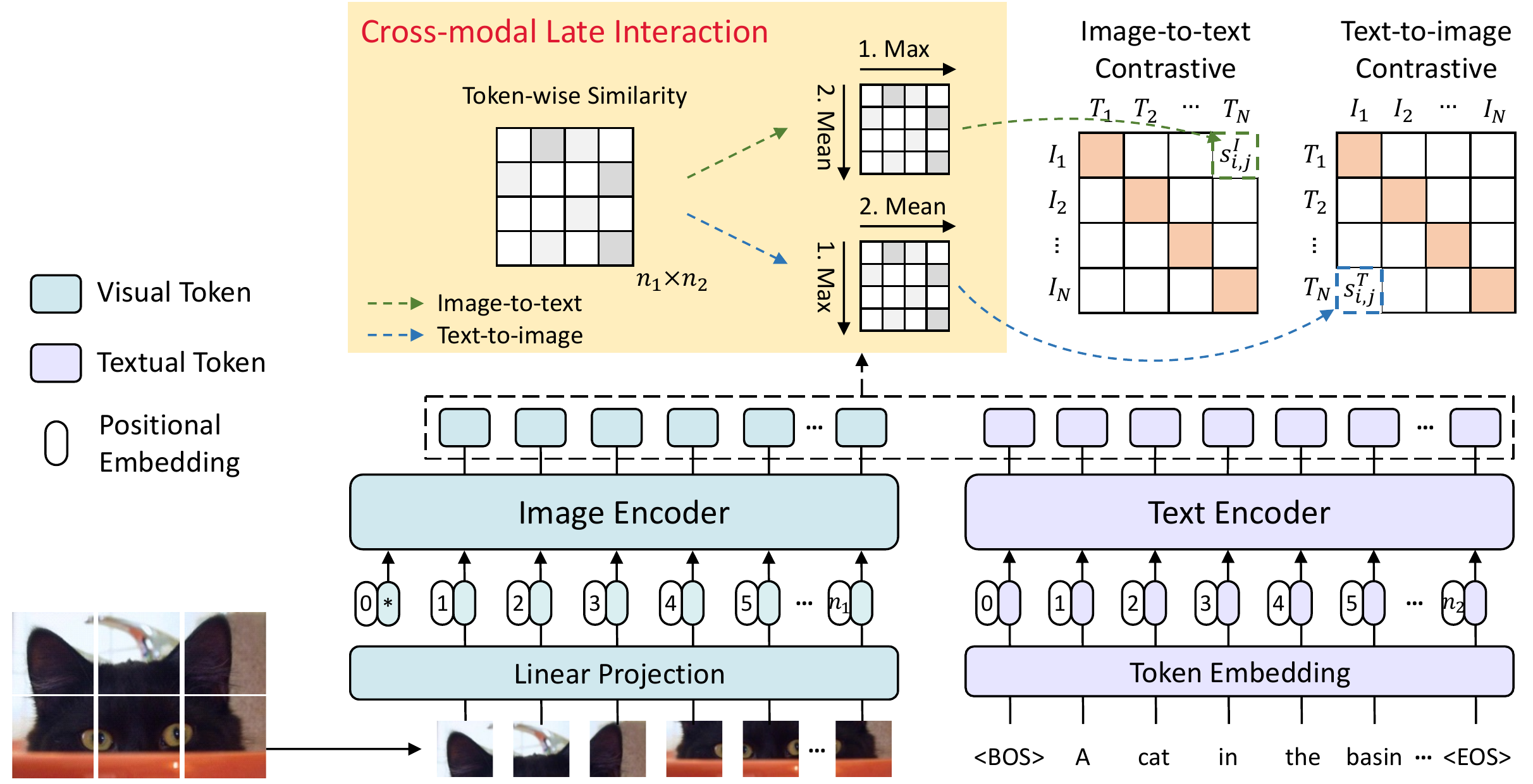}
\vspace{-6mm}
\caption{Overall architecture of FILIP, a dual-stream model with Transformer-based image and text encoders. On top of the image and text encoders,  the representations of textual tokens and visual tokens are linearly projected to the multi-modal joint space. A novel fine-grained contrastive learning equipped with cross-modal late interaction is proposed, which uses  a  token-wise  maximum  similarity  between visual and textual tokens. }
\label{fig:framework}
\end{center}
\vspace{-4mm}
\end{figure}

\label{sec:model_overview}

In this paper, we propose a new cross-modal pre-training model that excels in fine-grained interaction between image encoder and text encoder for mining more detailed semantic alignment, named as FILIP, as shown in  Figure \ref{fig:framework}. 
Particularly, FILIP is a dual-stream model with Transformer-based image and text encoders.
For the visual modality, the image encoder is a Vision Transformer \citep{dosovitskiy2020image} which takes the concatenation of an extra [CLS] token embedding and linearly projected image patches as input.
For the textual modality, following \cite{radford2021learning}, we use the lower-cased byte pair encoding (BPE) \citep{sennrich2016neural}
with a vocabulary size of 49,408  to tokenize the text. 
Each text sequence starts with [BOS] token and ends with [EOS] token.
After the word embedding layer, the token embeddings are fed into a modified decoder-only Transformer model as in \citep{radford2019language}. 
On top of the image and text encoders, the representations of textual tokens and visual tokens are linearly projected to the multi-modal common space, and are separately L2-normalized. 
Different from existing dual-stream models (e.g., CLIP and ALIGN) which models cross-modal interaction via only the global features of the entire image and text sequence, 
we introduce a novel fine-grained contrastive learning objective equipped with cross-modal late interaction which takes into account the fine-grained interaction between image patches and textual tokens, detailed in Section \ref{sec:late_interaction}.

\vspace{-1mm}
\subsection{Fine-grained Contrastive Learning}
\label{sec:late_interaction}

Contrastive representation learning has recently been found to learn better representations than its predictive counterpart 
in both visual \citep{tian2020contrastive} and vision-language cross-modal pre-training  \citep{radford2021learning}.
Under a general formulation of cross-modal contrastive learning \citep{radford2021learning}, we want to learn encoders $f_\theta$ for image data $\mathcal{I}$ and $g_\phi$ for text data $\mathcal{T}$ such that, given an image $ \xi \in \gI$, and a text $\xt \in \gT$,  the encoded representations $f_\theta(\xi)$ and $g_\phi(\xt)$ are close if they are related and far apart if not, under a distance metric. 
In each training batch, we sample $b$ image-text pairs $\{\xi_k, \xt_k\}_{k=1}^b$.
For image $\xi_k$ in image-text pair $\{\xi_k, \xt_k\}$, $\xt_k$ is its positive, while the other texts 
will be used as in-batch negatives. 
The image-to-text  contrastive loss $\gL^I_k$ for $\xi_k$  can then be formulated as
\[
\gL^I_k (\xi_k, \{\xt_j\}_{j=1}^b) = -\frac{1}{b} \log \frac{exp(s_{k,k}^I)}{\sum_{j} exp(s_{k,j}^I)},
\]
where $s_{k,j}^I$ denotes the similarity of the $k$-th image to the $j$-th text.
Similarly, the text-to-image contrastive loss for $\xt_k$ is
\[
\gL^T_k (\xt_k, \{\xi_j\}_{j=1}^b) = -\frac{1}{b} \log \frac{exp(s_{k,k}^T)}{\sum_{j} exp(s_{j,k}^T)}.
\]
The total loss of this mini-batch can be represented by 
\begin{equation}
\gL = \frac{1}{2}\sum\limits_{k=1}^b (\gL^I_k + \gL^T_k). \label{eq:contrastive_loss}
\end{equation}

\vspace{-1mm}
\subsubsection{Cross-modal Late Interaction}
\label{sec:Cross-modal-late}
From the contrastive loss (\ref{eq:contrastive_loss}), the cross-modal interaction is  reflected in how we compute the similarities $s_{i,j}^I$ and $s_{i,j}^T$ for the $i$-th image and $j$-th  text.
Previous methods like CLIP~\citep{radford2021learning} and ALIGN~\citep{jia2021scaling} simply encode each image or text separately to a global feature i.e., $f_\theta(\xi_i) \in \R^{d}$ and $g_\phi(\xt_j) \in \R^{d}$, and compute these two similarities as
\begin{equation}
   s^I_{i,j}  =   s^T_{i,j} =  f_\theta(\xi_i)^\top g_\phi(\xt_j), \label{eq:orig_loss}
\end{equation}
neglecting finer-grained interactions (e.g., word-patch alignment) between the two modalities.
To alleviate this problem, while simultaneously maintain the training and inference efficiency of dual-stream models, we apply a cross-modal late interaction inspired by \cite{khattab2020colbert} to model the token-wise cross-modal interaction.

Specifically, 
denote $n_1$ and $n_2$ as the  number of (non-padded) tokens of the $i$-th image  and $j$-th text, respectively, 
and the corresponding encoded features 
are $f_\theta(\xi_i) \in \R^{n_1 \times d}$ and $g_\phi(\xt_j) \in \R^{n_2 \times d}$.
For the $k$-th visual token, we compute its similarities with all textual tokens of $\xt_j$,  and use the largest one 
\begin{equation}
\max_{0\le r < n_2} [f_\theta(\xi_i)]_k^\top [g_\phi(\xt_j)]_r
\label{eq:tokenwise_max_sim}
\end{equation} 
as its  token-wise maximum similarity with $\xt_j$.
We then use the average token-wise maximum similarity of all non-padded tokens 
in the image (resp. text) as the similarity of an image to a text (resp. a text to an image). 
The similarity of the $i$-th image to the $j$-th text
can thus be formulated as:
\begin{equation}
s_{i,j}^I (\xi_i, \xt_j)  = \frac{1}{n_1}\sum_{k=1}^{n_1} [f_\theta(\xi_i)]_k^\top [g_\phi(\xt_j)]_{m_k^I}, \label{eq:late_sim_i}
\end{equation}
where $m_k^I = \arg \max_{0\le r < n_2} [f_\theta(\xi_i)]_k^\top [g_\phi(\xt_j)]_r$.
Similarly, the similarity of the $j$-th text to the $i$-th image is
\begin{equation}
s_{i,j}^T (\xi_i, \xt_j)  = \frac{1}{n_2}\sum_{k=1}^{n_2} [f_\theta(\xi_i)]_{m_k^T}^\top [g_\phi(\xt_j)]_k, \label{eq:late_sim_t}
\end{equation}
where $m_k^T = \arg \max_{0\le r < n_1} [f_\theta(\xi_i)]_r^\top [g_\phi(\xt_j)]_k$.
Note that $s_{i,j}^I (\xi_i, \xt_j)$ in   Equation (\ref{eq:late_sim_i}) does not necessarily equal $s_{i,j}^T (\xi_i, \xt_j)$ in Equation (\ref{eq:late_sim_t}).

\begin{remark}
\label{rmk:late_interaction_loss}
Intuitively, the token-wise maximum similarity in Equation~(\ref{eq:tokenwise_max_sim}) means that
for each image patch, we find its most similar textual token.
Similarly, for each textual token, we also find its closest image patch.
By applying this to the similarity calculation in (\ref{eq:late_sim_i}) and (\ref{eq:late_sim_t}) 
for contrastive loss (\ref{eq:contrastive_loss}),
 the dual-stream model learns fine-grained alignment between image patches and textual tokens.
\end{remark}

The original late interaction mechanism in \citep{khattab2020colbert} computes the  relevance score of a document to a query \textit{padded with mask tokens}, as a \textit{sum} of token-wise maximum similarities,
and is optimized via a \textit{pairwise} softmax cross-entropy loss.
Though inspired from \citet{khattab2020colbert},  our proposed cross-modal late interaction differs in several aspects.
Firstly, we exclude the padded textual tokens when computing the similarity, as they harm the performance. 
We speculate that this is because these padded tokens also learn textual representations and will mislead the model to align image patches to these meaningless padded tokens rather than meaningful non-padded words.
Secondly, when computing similarities (\ref{eq:late_sim_i}) and (\ref{eq:late_sim_t}), we use the average of the token-wise maximum similarities instead of summation in \citep{khattab2020colbert}. This is because the number of non-padded tokens varies from text to text, and this summation over all non-padded tokens can have quite different magnitudes, leading to less stabilized training and worse final performance.
Thirdly, we optimize the late interaction mechanism via a contrastive loss (\ref{eq:contrastive_loss}) which is found powerful  vision-language pre-training~\citep{radford2021learning} instead of the original pairwise loss in \citep{khattab2020colbert}.

\textbf{Training Efficiency.}
Though the cross-modal late interaction is able to capture finer-grained features 
compared with the original loss, 
it relies on the token-wise representations of both modalities, 
and can be inefficient in terms of communication, memory and computation, especially when the batch size is large. 
To alleviate this problem, we utilize several methods.
Firstly, we reduce the embedding size
to 256.
Besides, we reduce the precision of the last-layer features of both modalities from fp32 to fp16 before node communication in a distributed learning setting, 
and perform the multiplication in Equations (\ref{eq:late_sim_i}) and (\ref{eq:late_sim_t}) under the reduced precision.
In addition, since the complexity of similarity calculation scales with the sequence length of 
textual tokens and image patches,
for each image (resp. text), we select the  25\% tokens with the highest token-wise maximum similarity score (Equation (\ref{eq:tokenwise_max_sim})) among all texts (resp. images) in the same local worker before node communication, based on the intuition that each sample can be represented by a few of the most representative tokens.  Effects of these modifications are studied in Section \ref{sec:efficiency-study-of-late-loss}.
    

\subsubsection{Prompt Ensemble and Templates}
\label{sec:prompt_ensemble}

Due to the problem of polysemy and inconsistency with the pre-training process, following \citet{radford2021learning}, we also use prompt templates to augment the original label for some downstream tasks.
For visualizations, for simplicity, we 
use only one prompt template across the paper, i.e. ``a photo of a \{label\}.'' as  \citet{radford2021learning}.
For other experiments, we
report results using prompt ensemble following~\citet{radford2021learning}.
When multiple prompts are allowed, the token-wise representations of different prompt templates for the same class label are different, and can not be summed together to form 
a mean textual representation as in \citep{radford2021learning}.
Thus, instead of ensembling different prompt templates by their mean textual representation, we ensemble them
by their mean token-wise similarity.
Specifically, suppose there are $C$ prompt templates, each label is augmented to $C$ different texts $\xt_1, \xt_2, \cdots, \xt_C$.
The 
similarity between an image $\xi$ and this label is computed as
$
\frac{1}{C}\sum_{c=1}^C s_{\cdot,\cdot}^I (\xi, \xt_c),
$
where $s_{\cdot,\cdot}^I$ is defined in Equation (\ref{eq:late_sim_i}).

We use a unified rule-based method inspired by \citet{radford2018improving}
to construct  prompt templates for image classification tasks.
Specifically, each template consists of four components:

\vspace{-3mm}
\begin{equation}
\text{[prefix] \{label\}, [category description]. [suffix].} \label{eq:prompt_template}
\end{equation}
Here, the ``[prefix]'' is an in-context description like ``a photo of a" similar as~\cite{radford2021learning};
``{label}'' is a class label of the dataset;
``[category description]'' describes the category which is found helpful for some fine-grained image classification datasets \citep{radford2021learning}, e.g.,  `` a type of pet'' for dataset Oxford-IIIT Pets.
An interesting finding is that,
adding a suffix that includes the reference word ``it" (e.g., ``I like it.") at the end of the prompt empirically improves the zero-shot classification performance of the proposed model.
We speculate this is because the reference word ``it" strengthens the fine-grained cross-modal alignment, as it
can also be aligned to image patches of the target object. Detailed prompt templates  for different datasets can be found in Appendix~\ref{apdx:prompt_template}.

\vspace{-1mm}
\subsection{Image and Text Augmentation}

\label{sec:augmentation}
To obtain better generalization and data-efficiency of the model, we perform data augmentation on both images and texts during the pre-training phase to construct more image-text pairs. 
We apply AutoAugment \citep{krizhevsky2012imagenet,sato2015apac,cubuk2019autoaugment,hoffer2020augment} for image augmentation, following the SOTA vision recognition methods \citep{touvron2021training,xie2020self}.
To ensure the augmented texts are semantically similar as the original one, for
text augmentation, we rewrite the original text using 
back-translation \citep{xie2020unsupervised,sennrich2016improving}.
Specifically,
the texts are first translated to the target language and then translated back to the source language. 
We choose German and Russian as the target language and get extra two texts for each image-text pair. 
When constructing a batch of image-text pairs during the pre-training, the text of each image-text pair is randomly sampled from the three candidate texts, i.e., the original text and two back-translated texts.

\vspace{-1mm}
\subsection{Pre-training Dataset}
\label{sec:dataset_construction}

A sufficiently large image-text dataset is a prerequisite for vision-language pre-training. 
Recent CLIP \citep{radford2021learning} and ALIGN \citep{jia2021scaling} construct datasets with 400M and 1800M image-text pairs, respectively. 
In this work, we also construct a large-scale dataset called FILIP300M, which consists of 300M image-text pairs and covers board vision and language concepts.
Specifically, we collect image-text pairs from the Internet,
and apply the following image- and text-based filtering rules to clean data.
For image-based filtering, we remove the images whose shorter dimension is smaller than 200 pixels and the aspect ratio is larger than 3. 
For text-based filtering, we keep only English texts, and exclude the meaningless ones, e.g., img\_0.jpg. 
We also discard image-text pairs whose texts are repeated for over 10 times. 
Besides, we also use 3 public datasets, including Conceptual Captions 3M (CC3M) \citep{sharma2018conceptual}, Conceptual 12M (CC12M) \citep{changpinyo2021cc12m} and Yahoo Flickr Creative Commons 100M (YFCC100M) \citep{thomee2016yfcc100m}. We apply the same filtering rules on YFCC100M. Finally, we use about 340M image-text pairs for pre-training. 
Despite using a smaller training dataset than CLIP and ALIGN, our models still outperform them in most down-steam tasks (see Section~\ref{sec:expt}).

\begin{table}
\vspace{-2mm}
\Large
\caption{Top-1 accuracy(\%) of zero-shot image classification on 12 datasets. Our FILIP can boost 3$\sim$5\% accuracy on average.}
\label{zeroshot-classification-table}
\vspace{-1mm}
\begin{center}
\resizebox{\textwidth}{!}{
\begin{tabular}{l|cccc cccc cccc |c}

&\rotatebox{90}{\Large{CIFAR10}}~~ &
\rotatebox{90}{\Large{CIFAR100}}~~ &
\rotatebox{90}{\Large{Caltech101}}~~ &
\rotatebox{90}{\Large{StanfordCars}}~~ &
\rotatebox{90}{\Large{Flowers102}}~~ &
\rotatebox{90}{\Large{Food101}}~~ &
\rotatebox{90}{\Large{SUN397}}~~ &
\rotatebox{90}{\Large{DTD}}~ &
\rotatebox{90}{\Large{Aircrafts}}~~ &
\rotatebox{90}{\Large{OxfordPets}}~~ &
\rotatebox{90}{\Large{EuroSAT}}~~ & 
\rotatebox{90}{\Large{\textbf{ImageNet}}}~~ &
\rotatebox{90}{\Large{\textbf{Average}}}~~ \\
\midrule
CLIP-ViT-B/32 & 91.3 & 65.1  & 87.9 & 59.4 & 66.7 & 84.4 & 63.2 & 44.5 & 21.2 & 87.0 & 49.4 & 63.2 & 65.3 \\
$\text{FILIP}_{\text{base}}$-ViT-B/32 & 86.9 & 65.5  & 91.9 & 55.4 & 85.3 & 82.8 & 69.1 & 49.3 & 57.2 & 88.1 & 49.9 & 68.8 & \textbf{70.9}$^{+5.6}$ \\

\midrule
CLIP-ViT-L/14 &  96.2 & 77.9 & 92.6 & 77.3 & 78.7 & 92.9 & 67.7 & 55.3 & 36.1 & 93.5 & 59.9 & 75.3 & 75.3 \\ 
$\text{FILIP}_{\text{large}}$-ViT-L/14 & 95.7 & 75.3  & 93.0 & 70.8 & 90.1 & 92.2 & 73.1 & 60.7 & 60.2 & 92 & 59.2 & 77.1 & \textbf{78.3}$^{+3.0}$ \\

\bottomrule
\end{tabular}}
\end{center}
\vspace{-2mm}
\end{table}

\vspace{-1mm}
\section{Experiments}
\label{sec:expt}

\vspace{-1mm}
\subsection{Experimental Setup}
\label{sec:experiment_details}
\textbf{Model Architectures.}
We train two 
models from scratch, i.e., $\text{FILIP}_{\text{base}}$ and $\text{FILIP}_{\text{large}}$. 
The model architectures follow
CLIP \citep{radford2021learning}, i.e., the image encoder is ViT-B/32 for $\text{FILIP}_{\text{base}}$ and ViT-L/14 for $\text{FILIP}_{\text{large}}$. More details can be found in Appendix \ref{apdx:expt_setting}.

\textbf{Pre-training Details.}
To save memory and scale up the batch size, automatic mixed-precision \citep{micikevicius2018mixed} and gradient checkpoint \citep{griewank2000algorithm, chen2016training} are used
The input images are resized to $224 \times 224$ resolution during pre-training and the maximum length of the text is limited to $77$ tokens following \citet{radford2021learning}.
The training is mainly conducted on Nvidia V100 GPUs and Ascend Cards.
$\text{FILIP}_{\text{base}}$ is trained on 128 cards about 9 days and $\text{FILIP}_{\text{large}}$ takes about 24 days to train on 192 cards. 
Unless otherwise specified, we use $\text{FILIP}_{\text{large}}$ to compare with other methods and $\text{FILIP}_{\text{base}}$ for ablation.
We train both models using the LAMB optimizer \citep{2019Large} and cosine learning rate schedule \citep{2016SGDR} with a linear warmup. 
Weight decay regularization is applied to all parameters except bias, layer normalization, token embedding, positional embedding and temperature in contrastive loss. 
Detailed values of hyperparameters for different datasets and models can be found in Appendix \ref{apdx:expt_setting}.


\vspace{-1mm}
\subsection{Zero-Shot Image Classification}
\label{sec:zeroshot_classification}
In this section, we evaluate our proposed FILIP on the zero-shot image classification task.
We compare our FILIP with CLIP \citep{radford2021learning} on 12 downstream classification datasets, using the same evaluation setting as in CLIP. As described in Section \ref{sec:prompt_ensemble}, we  apply a set of prompts for each dataset and ensemble them to get the final results, see Appendix \ref{apdx:prompt_template} for details. We only compare the zero-shot performance with CLIP here as ALIGN does not release its model and  the related performances are not reported in their paper.

Table \ref{zeroshot-classification-table} shows the results on 12 datasets.
Despite using less training data (340M vs. 400M), both $\text{FILIP}_{\text{base}}$ and $\text{FILIP}_{\text{large}}$ considerably outperform their CLIP counterparts in terms of average top-1 accuracy over 12 datasets, i.e., achieving  absolute improvements of 5.6\% and 3.0\%, respectively. In particular, our FILIP surpasses CLIP on  ImageNet, the largest dataset among 12 datasets.
FILIP also achieves substantial performance gains on some domain-specific datasets, e.g., for Aircrafts, the two FILIP models reach a 30\% improvement over CLIP on average. We speculate this is because,
unlike CLIP which aggregates the information of the whole image into the representation of the [CLS] token, our proposed FILIP model focuses more on the target object by directly aligning the image patches corresponding to the target object with the textual tokens corresponding to the class label (visualizations of word-patch alignment are in Section \ref{sec:Visualization of Fine-grained Alignment}).



\begin{table}
\vspace{-3mm}
\center
\caption{Results of zero-shot image-text retrieval on Flickr30K and MSCOCO datasets. The last two rows (marked with *) report the zero-shot results on Flickr30K dataset of model fine-tuned on MSCOCO dataset, following the setting of ALBEF \citep{li2021align}.}
\huge
\label{tab:zero-shot-retrieval-table}
\resizebox{\textwidth}{!}{
\begin{tabular}{ccccccccccccc}
\toprule
            & \multicolumn{6}{c}{Flickr30K}                                                               & \multicolumn{6}{c}{MSCOCO}                                                                  \\
            & \multicolumn{3}{c}{image-to-text} & \multicolumn{3}{c}{text-to-image} & \multicolumn{3}{c}{image-to-text} & \multicolumn{3}{c}{text-to-image} \\
            & R@1           & R@5           & R@10         & R@1           & R@5           & R@10         & R@1           & R@5           & R@10         & R@1           & R@5           & R@10         \\
\midrule
Unicoder-VL & 64.3          & 85.8          & 92.3         & 48.4          & 76.0            & 85.2         & $-$             & $-$             & $-$            & $-$             & $-$             & $-$            \\
ImageBERT   & 70.7          & 90.2          & 94.0           & 54.3          & 79.6          & 87.5         & 44.0            & 71.2          & 80.4         & 32.3          & 59.0            & 70.2         \\
UNITER      & 83.6          & 95.7          & 97.7         & 68.7          & 89.2          & 93.9         & $-$             & $-$             & $-$            & $-$             & $-$             & $-$            \\
CLIP        & 88.0            & 98.7          & 99.4         & 68.7          & 90.6          & 95.2         & 58.4          & 81.5          & 88.1         & 37.8          & 62.4          & 72.2         \\
ALIGN       & 88.6          & 98.7          & 99.7         & \textbf{75.7}          & \textbf{93.8}          & \textbf{96.8}         & 58.6          & 83.0            & 89.7         & 45.6          & 69.8          & 78.6         \\
\textbf{FILIP}      & \textbf{89.8}          & \textbf{99.2}          & \textbf{99.8}         & {75.0}            & {93.4}          & {96.3}         & \textbf{61.3}          & \textbf{84.3}          & \textbf{90.4}         & \textbf{45.9}          & \textbf{70.6}          & \textbf{79.3}         \\ \hline
ALBEF*       & 94.1          & 99.5          & 99.7         & 82.8          & 96.3          & 98.1         & $-$             & $-$             & $-$            & $-$             & $-$             & $-$            \\
\textbf{FILIP}*       & \textbf{95.4}          & \textbf{99.8}          & \textbf{100.0}         & \textbf{84.7}            & \textbf{97.0}          & \textbf{98.7}         & $-$             & $-$             & $-$            & $-$             & $-$             & $-$            \\
\bottomrule         
\end{tabular}}
\vspace{-3mm}
\end{table}

\vspace{-1mm}
\subsection{Image-Text Retrieval}
\label{sec:image_text_retrieval}
Image-text retrieval consists of two sub-tasks: image-to-text retrieval and text-to-image retrieval. 
We evaluate our FILIP model on two retrieval benchmark datasets: Flickr30K \citep{dataset_flickr30k} and MSCOCO \citep{dataset_mscoco}, under both zero-shot and fine-tuned settings. 
More details of experimental setting can be found in Appendix \ref{apdx:expt_setting}.

Tables \ref{tab:zero-shot-retrieval-table} and  \ref{tab:finetuned-retrieval-table} show the results of 
zero-shot and fine-tuned 
image-text retrieval,
respectively. We compare our FILIP model against methods with complex attention layers including Unicoder-VL \citep{unicoder_vl}, ImageBERT \citep{imagebert}, UNITER \citep{chen2020uniter}, VILLA \citep{villa}, ERNIE-ViL \citep{ernie_vil}, Oscar \citep{li2020oscar}, VinVL \citep{zhang2021vinvl}, ALBEF \citep{li2021align}, and methods trained on larger-scale image-text datasets including CLIP \citep{radford2021learning} and ALIGN \citep{jia2021scaling}. As we can see, FILIP achieves state-of-the-art performances under all metrics on both Flickr30K and MSCOCO datasets, except for zero-shot text-to-image retrieval on Flickr30K, where FILIP achieves competitive performance with SOTA. For zero-shot image-to-text retrieval on MSCOCO dataset, the absolute R@1 of our proposed FILIP is 2.7\% higher than ALIGN, which is trained on a much larger dataset.

\begin{table}
\vspace{-3mm}
\center
\caption{Results of 
fine-tuned 
image-text retrieval on Flickr30K and MSCOCO datasets.}
\Large
\label{tab:finetuned-retrieval-table}
\resizebox{\textwidth}{!}{
\begin{tabular}{ccccccccccccc}
\toprule
            & \multicolumn{6}{c}{Flickr30K}                                                               & \multicolumn{6}{c}{MSCOCO}                                                                  \\
            & \multicolumn{3}{c}{image-to-text} & \multicolumn{3}{c}{text-to-image} & \multicolumn{3}{c}{image-to-text} & \multicolumn{3}{c}{text-to-image} \\
            & R@1           & R@5           & R@10         & R@1           & R@5           & R@10         & R@1           & R@5           & R@10         & R@1           & R@5           & R@10         \\
\midrule
Unicoder-VL & 86.2          & 96.3          & 99.0           & 71.5          & 90.9          & 94.9         & 62.3          & 87.1          & 92.8         & 48.4          & 76.7          & 85.9         \\
ImageBERT   & 87.0            & 97.6          & 99.2         & 73.1          & 92.6          & 96.0           & 66.4          & 89.8          & 94.4         & 50.5          & 78.7          & 87.1         \\
UNITER      & 87.3          & 98.0            & 99.2         & 75.6          & 94.1          & 96.8         & 65.7          & 88.6          & 93.8         & 52.9          & 79.9          & 88.0           \\
VILLA       & 87.9          & 97.5          & 98.8         & 76.3          & 94.2          & 96.8         & $-$             & $-$             & $-$            & $-$             & $-$             & $-$            \\
ERNIE-ViL   & 88.1          & 98.0            & 99.2         & 76.7          & 93.6          & 96.4         & $-$             & $-$             & $-$            & $-$             & $-$             & $-$            \\
Oscar       & $-$             & $-$             & $-$            & $-$             & $-$             & $-$            & 73.5          & 92.2          & 96.0           & 57.5          & 82.8          & 89.8         \\
VinVL       & $-$             & $-$             & $-$            & $-$             & $-$             & $-$            & 75.4          & 92.9          & 96.2         & 58.8          & 83.5          & 90.3         \\
ALIGN       & 95.3          & 99.8          & 100.0          & 84.9          & 97.4          & 98.6         & 77.0            & 93.5          & 96.9         & 59.9          & 83.3          & 89.8         \\
ALBEF       & 95.9          & 99.8          & \textbf{100.0}          & 85.6          & 97.5          & 98.9         & 77.6          & 94.3          & 97.2         & 60.7          & \textbf{84.3}          & 90.5         \\
Our FILIP       & \textbf{96.6}          & \textbf{100.0}          & \textbf{100.0}          & \textbf{87.1}          & \textbf{97.7}          & \textbf{99.1}         & \textbf{78.9}          & \textbf{94.4}          & \textbf{97.4}         & \textbf{61.2}          & \textbf{84.3}          & \textbf{90.6}        \\
\bottomrule
\end{tabular}}
\vspace{-3mm}
\end{table}

\begin{table}
\vspace{-3mm}
\caption{Ablation study of different components on pre-training subset of YFCC100M. I2T and T2I are abbreviations for image-to-text and text-to-image retrieval, respectively. ``ZS'' means zero-shot performance. Underlined numbers have
the highest improvements for the corresponding metrics. }
\label{ablation-yfcc-table}
\begin{center}
\begin{tabular}{l|rrrrr}
\toprule \multirow{2}{*}{ Model } & \multicolumn{4}{c} { MSCOCO }         & ImageNet \\
                                 & I2T R@1 & I2T R@5 & T2I R@1 & T2I R@5 & ZS Top1 \\
\midrule Baseline (ViT-B/32) & $25.0$ & $49.5$ & $14.7$ & $34.7$ & 30.4 \\ 
~w/ image augmentation & $26.1$ & $51.8$ & $16.5$ & $37.5$ & $ 32.5 $ \\  
~w/ back translation & $29.2$ & $55.0$ & $17.9$ & $39.8$ & $33.9$ \\ 
~w/ cross-modal late interaction & $\underline{30.5}$ & $\underline{55.3}$ & $\underline{18.5}$ & $\underline{40.0}$ & $\underline{34.3}$\\
Our $\text{FILIP}_{\text{base}}$ & $\mathbf{33.4}$ & $\mathbf{60.1}$ & $\mathbf{23.0}$ & $\mathbf{46.2}$ & $\mathbf{37.8}$ \\  
\bottomrule
\end{tabular}
\end{center}
\vspace{-3mm}
\end{table}

\begin{table}[t!]
\vspace{-3mm}
    \caption{Efficiency study of the cross-modal late interaction. ``orig'' and ``late'' stand for the contrastive loss based on the original cosine similarity 
    in CLIP and our proposed cross-modal late interaction, respectively. ``ZS'' means zero-shot performance.
    We report results for ViT-B/32 trained on filtered YFCC100M with 8 V100 GPUs, with a batch size of 512 per GPU. Training time and memory consumption are tested using the same gradient checkpoint configuration. 
    * denotes our final setting used in other experiments.}
    \label{tab:efficiency-late}
    \begin{center}

    \begin{tabular}{cccccccc}
    \toprule \multirow{2}{*}{ Loss }   &Embed&Embed &  Token  &  Training time   &  Memory  & ImageNet \\
                                     & dim & precision&  \% & (sec/iter)     &  (MB)    & ZS Top1  \\
    \midrule 
    orig (baseline)& 512 & fp32 & - &1.31 & 14300  & 30.4 \\
    late & 512 & fp32 &100\% & 2.85 & 26000 & 34.6 \\
    late & 512 & fp16 &100\% & 2.67 & 23468 & 34.5 \\
    late & 256 & fp16 &100\% & 2.31 & 22382 & \textbf{35.2} \\
    late & 256 & fp16 &50\% & 1.61 & 16336 & 34.5 \\
    late* & 256 & fp16 &25\% & 1.39 & 16100 & 34.3 \\
    
    \bottomrule
    \end{tabular}
    \end{center}
    \vspace{-3mm}
\end{table}

\vspace{-1mm}
\subsection{Ablation Study}
\label{sec:ablation_yfcc}

\textbf{Effectiveness of Each Component.}
We study the effectiveness of each component  in FILIP, i.e., image/text augmentations and cross-modal late interaction. Experiments are conducted on 
$\text{FILIP}_{\text{base}}$,
with a filtered subset of YFCC100M as the training dataset (as described in Section \ref{sec:dataset_construction}),
on both zero-shot retrieval and classification tasks. 
We measure models' performance on  MSCOCO zero-shot image-text retrieval and ImageNet zero-shot classification, which are two effective indicators 
for the quality of the learned vision-language representations. 

Table \ref{ablation-yfcc-table} reports the results. As can be seen, all three components 
are beneficial for both tasks.
Despite the simple design, cross-modal late interaction brings significant performance improvements over the baseline (the vanilla CLIP ViT-B/32), with an absolute R@1 gain of 5.5\% (resp. 3.8\%)  for image-to-text (resp. text-to-image) retrieval on MSCOCO and an absolute top-1 accuracy gain of 3.9\% for zero-shot classification on ImageNet. 
Further improvements are observed when all components are combined together.



\begin{figure}
\vspace{-3mm}
	\centering
	\includegraphics[width=\linewidth]{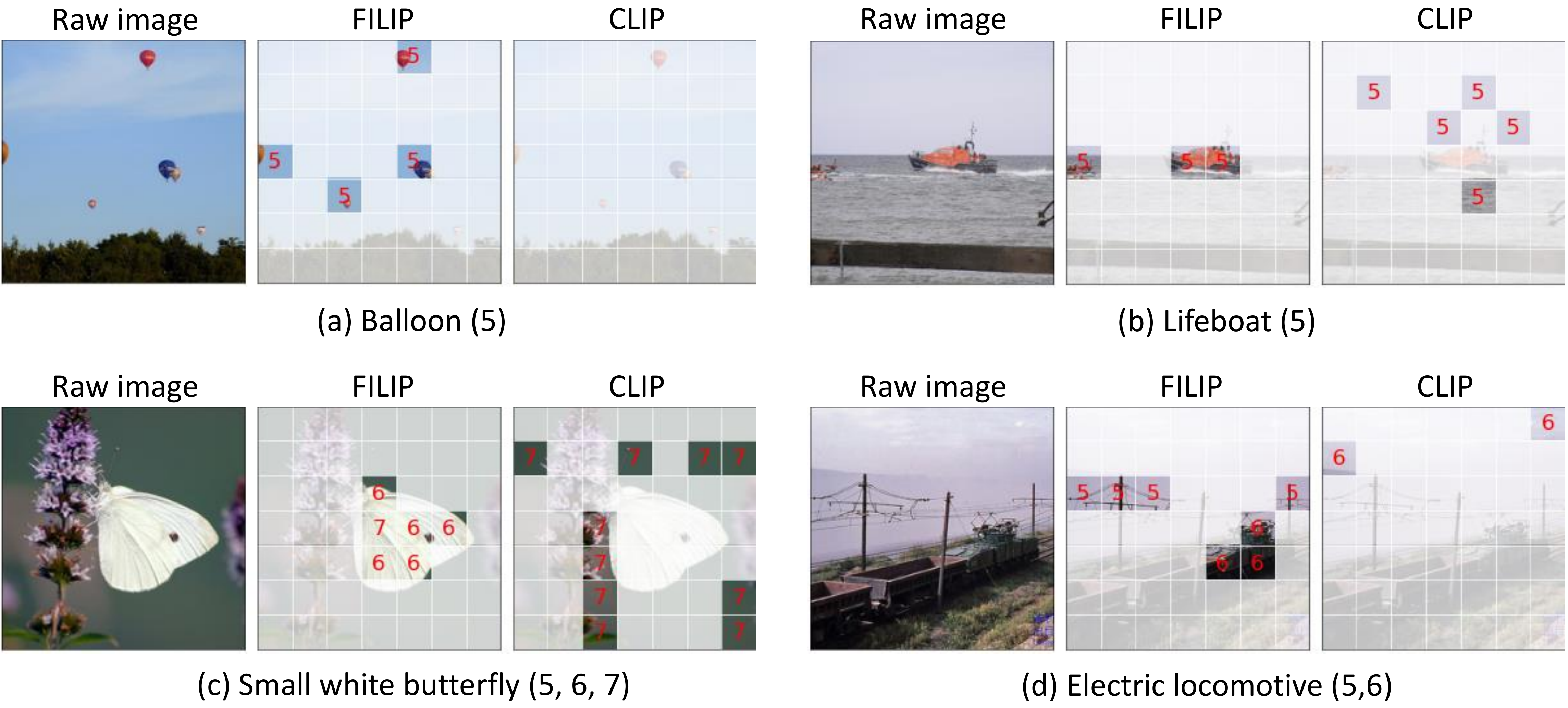}
	\vspace{-5mm}
	\caption{Visualizations of word-patch alignment for 4 classes of the ImageNet dataset and ``a photo of a \{label\}." is the prompt. Numbers in the parentheses after the class label indicate the location indices of the class label in the tokenized textual sequence. The correct predictions are highlighted by opaque patches with the class label indices in red.}
	\label{fig:single_obj}
	\vspace{-3mm}
\end{figure}

\textbf{Efficiency Study of Cross-modal Late Interaction.}
\label{sec:efficiency-study-of-late-loss}
Since the late interaction mechanism in Section~\ref{sec:Cross-modal-late} requires to calculate the similarity between all visual and textual tokens, 
its efficiency can be a problem when
employed in large-scale distributed training.
As described in Section \ref{sec:Cross-modal-late}, we make several attempts to address the issue. 
Table \ref{tab:efficiency-late} shows the efficiency improvement on zero-shot classification on ImageNet when
these attempts are applied. 
As can be seen, these attempts improve the efficiency of late interaction  
without accuracy drop. Combining all three attempts achieves only slightly slower training and larger memory consumption than the original loss in CLIP.


\vspace{-1mm}
\subsection{Visualization of Fine-grained Alignment}
\label{sec:Visualization of Fine-grained Alignment}

In this section, we visualize FILIP's capability of capturing fine-grained cross-modal correspondence using the method of word-patch alignment. To make a fair comparison, we use our $\text{FILIP}_{\text{base}}$ trained on YFCC100M and CLIP's ViT-B/32, which are of the same size, for visualization. Each image is patchified to $7\times 7$ image patches. 
More visualization results can be found in Appendix~\ref{apdx:more_vis}.

\textbf{Visualization Method.} The word-patch alignment is performed based on the token-wise similarity between the image patches and textual tokens. Specifically, for the $k$-th image patch, the location index of textual token   with the largest similarity with it ($m_k^I$ in Equation~(\ref{eq:late_sim_i})) is considered as its predicted label, and is placed at the center of it.
Take class ``balloon'' as an example. 
There are 8 tokens in the tokenized textual sequence ``[BOS] a photo of a balloon. [EOS]'', 
and the location index of the class label ``balloon'' is ``5''. 
Note that one class label may be tokenized to more than one token.
Location indices of textual tokens corresponding to the class label are highlighted in red, while the others are marked in white.
A desired model that learns fine-grained representations  would predict  image patches of the target object to red indices.

\textbf{Observations.}
Figure~\ref{fig:single_obj} shows the word-patch alignment results for FILIP and CLIP on 4 classes from the ImageNet dataset.
As can be seen,
FILIP exhibits the finer-grained understanding of 
an image in the following aspects. 
(i) A single object: 
From the visualization of class ``small white butterfly'', the image patches covering the object are all classified correctly;
(ii) Same object in different shapes:
From the visualizations of class ``balloon'' and ``lifeboat'', image patches corresponding to all target objects with different shapes and locations  are correctly classified; 
(iii) Key Components of an object: For class ``electric locomotive'', there are two key components crucial to correctly classifying the image, i.e., ``electric'' and ``locomotive'', whose corresponding textual token indices are ``5'' and ``6'', respectively. As can be seen, image patches matching these two key components are respectively correctly classified.
On the other hand, CLIP can not correctly align image patches with corresponding textual tokens.
Compared with \cite{kim2021vilt} which uses an extra optimal transport to align the textual word and image patch distributions, the word-patch alignment can be simply automatically learned by our method.

\vspace{-2mm}
\section{Conclusion and Future Work}
\label{sec:conclusion}
This paper introduces FILIP, a simple yet generic framework towards fine-grained vision-language pre-training.
By using a token-wise maximum similarity, our method learns fine-grained representation for patches in the images and words in the sentences.
While it achieves competitive results against several large-scale multi-modal pre-training on various downstream tasks, both its architecture and training procedure can still be optimized to improve its performance. In the future, a
more advanced image encoder as well as a well-designed interaction layer can be used to boost the performance.
Furthermore, we can further add more masked language/image loss to support more generation tasks.
To this end, we hope to extend FILIP as a generic and unified interface for solving a large variety of vision-language tasks.

\bibliography{iclr2022_conference}
\bibliographystyle{iclr2022_conference}
\newpage
\appendix
\section{Appendix}

\subsection{Datasets Summary}
Table \ref{tab:dataset_statistics} shows the number of image-text pairs of each datasets used in different pre-training methods.
\begin{table}[h]
\center
\caption{Number of image-text pairs used in the pre-training of FILIP, CLIP and ALIGN.}
\label{tab:dataset_statistics}
\begin{tabular}{ c|cccc|c|c } 
 \hline
\multirow{2}{*}{} & \multicolumn{4}{c|} { FILIP } & { CLIP } & { ALIGN } \\
& CC3M & CC12M & YFCC100M & FILIP300M & \citep{radford2021learning} & \citep{jia2021scaling} \\ 
  \hline
 \# & 3M & 10M & 26M & 300M & 400M & 1800M \\
 \hline
\end{tabular}
\end{table}

\subsection{Detailed Experimental Settings}
\label{apdx:expt_setting}

\begin{table}[htbp]
\caption{The architecture parameters for FILIP models.}
\label{tab:Filip_model_Hyperparameter}
    \centering
    \begin{tabular}{l|cc ccc ccc}
        \toprule 
        \multirow{2}{*}{ Model } & Embedding & Input & \multicolumn{3}{c}{ Image Encoder } & \multicolumn{3}{c}{ Text Encoder } \\
          & dimension & resolution & \#layers & width & \#heads & \#layers & width & \#heads \\
        \midrule 
        
        $\text{FILIP}_{\text{base}}$  &  256 & $224\times 224$ & 12 & 768 & 12 & 12 & 512 & 8 \\
        $\text{FILIP}_{\text{large}}$ &  256 & $224\times 224$ & 24 & 1024 & 16 & 12 & 768 & 12 \\
        \bottomrule
    \end{tabular}
\end{table}

\paragraph{Model Architectures.} 
We follow the same architecture design as CLIP, for both $\text{FILIP}_{\text{base}}$ and $\text{FILIP}_{\text{large}}$, except that we reduce the embedding dimension from 512/768 to 256 for the efficiency of loss computation.
Table \ref{tab:Filip_model_Hyperparameter} describes the details of architectures.

\begin{table}
    \centering
    \caption{Common hyperparameters used for FILIP pre-training.}
    \label{tab:pre-training hyperparams}
    \centering
     \begin{tabular}{l|c}
        \toprule Hyperparameter & Value \\
        \midrule
        Vocabulary size & 49408 \\
        Initial temperature & $0.07$ \\
        LAMB $\beta_{1}$ & $0.9$ \\
        LAMB $\beta_{2}$ & $0.999$ \\
        LAMB $\epsilon$  & $10^{-4}$ \\
        Warm-up iters & 3000 \\
        Training epochs & 30  \\
        \bottomrule
    \end{tabular}
 
    \end{table}
  
\begin{table}
    \centering
    \caption{Model- and dataset-specific hyperparameters used for FILIP pre-training.
    Numbers in batch size represent the total batch size across all workers and are calculated as: batch size per GPU $\times$ \#GPUs. FILIP340M is the combination of FILIP300M, YFCC100M, CC12M and CC3M.}
    \label{tab:Model_and_dataset_specific hyperparameters}
    \begin{minipage}{\textwidth}
        \centering
          \begin{tabular}{l|l|cccc}
        \toprule 
        Model  & Dataset  & Batch size & Base LR & Weight decay &  \\
        \midrule 
        $\text{FILIP}_{\text{base}}$ & YFCC100M & $1024 \times 8 $  & $6 \times 10^{-3}$ & 3e-2 \\
        $\text{FILIP}_{\text{base}}$  & FILIP340M & $ 320 \times 128 $ & $ 2 \times 10^{-3}$ & 3e-3  \\
        \midrule
        $\text{FILIP}_{\text{large}}$ & FILIP340M & $ 160 \times 192 $ &  $ 8 \times 10^{-4}$ & 3e-3  \\ 
        
        \bottomrule
    \end{tabular}
    \end{minipage}
\end{table}

\paragraph{Details for Pre-training and Hyperparameters.}  
For the implementation of the contrastive loss, following CLIP \citep{radford2021learning} and ALIGN \citep{jia2021scaling}, we also set the temperature in the softmax function to be a learnable parameter and initialize it as 0.07. 
For the pre-training, we use the LAMB optimizer implemented by the cybertronai's open-source repository (\url{https://github.com/cybertronai/pytorch-lamb}). 
For the learning rate scheduler, we first assign a base learning rate and then linearly warm it up to the peak learning rate according to the effective total batch size by a square root strategy, $peak\_lr=base\_lr \times \sqrt{\frac{total\_bs}{512}}$. 
We note that a large weight decay is crucial to stabilize training and improve generalization.
Specifically, we found that the training stability is a challenging issue when applying mix-precision training to large-scale models, i.e.,  the training is extremely unstable and the NaN loss easily happens. Recent works DALL-E \citep{ramesh2021zeroshot} and Cogview \citep{ding2021cogview} also notice this issue and provide their solutions. 
However, we found that simply increasing the weight decay and applying the trick of removing the weight decay of specific parameters as described in Section \ref{sec:experiment_details} work for our case.
The base learning rate and weight decay are selected manually via observing the performance at the early training stage.
Table \ref{tab:pre-training hyperparams} summarizes the common hyperparameters  and Table \ref{tab:Model_and_dataset_specific hyperparameters} shows the model- and dataset-specific hyperparameters for FILIP pre-training.



\paragraph{Details for Image-text Retrieval.} Following previous works \citep{jia2021scaling,li2021align}, for Flickr30K, we test on the 1K test set with or without fine-tuning on the 30K training set, while for MSCOCO, we test on the 5K test set with or without fine-tuning on the 113K training set.  We use the similarity between image and text for ranking and use the contrastive loss for fine-tuning. Since there are multiple texts for each image in these two datasets, we change the ground-truth label of contrastive loss to consider multiple positives, by assigning a probability of 1/\#positive to each positive following ALBEF~\citep{li2021align}. Besides, we also use prompts during evaluation for both datasets, see Appendix \ref{apdx:prompt_template} for details. Table \ref{tab:hyperparameter_image_text_retrieval} shows the hyperparameters for image-text retrieval fine-tuning.

\begin{table}[htbp]
    \centering
    \caption{Hyperparameters used for image-text retrieval fine-tuning.}
    \label{tab:hyperparameter_image_text_retrieval}
    \begin{tabular}{l|c}
        \toprule Hyperparameter & Value \\
        \midrule
        Image size & 392 $\times$ 392 \\
        Training epochs & 3 \\
        Optimizer & LAMB \\ 
        Batch size & 5120 \\
        Base LR & $2 \times 10^{-4}$ \\
        Weight decay & $3 \times 10^{-4}$ \\
        \bottomrule
    \end{tabular}
\end{table}

\subsection{More visualizations of Word-patch Alignment and Grad-cam Heatmaps}
\label{apdx:more_vis}
\begin{figure}
\begin{center}
\includegraphics[width=\textwidth]{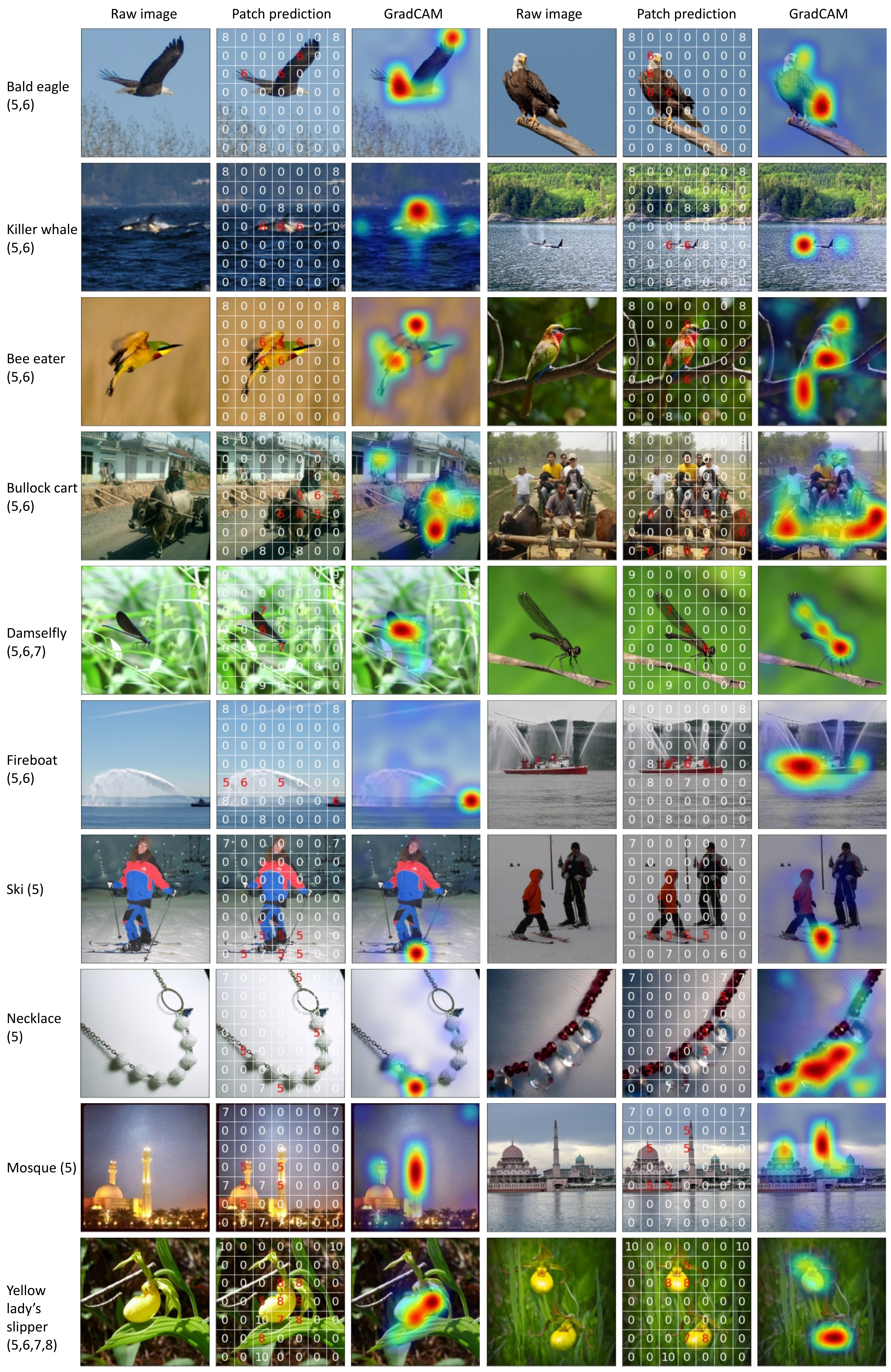}
\caption{More visualizations on different classes of ImageNet dataset. Numbers in the parentheses after the class label indicate the location indices of class label in the tokenized textual sequence. }
\label{fig:MoreImageNetVis}
\end{center}
\end{figure}

In Figure~\ref{fig:MoreImageNetVis}, we visualize the cross-modal alignment of the proposed method for more images, in terms of
both word-patch alignment as described in Section~\ref{sec:Visualization of Fine-grained Alignment} and Grad-CAM heatmaps~\citep{selvaraju2017grad}.
We compute the Grad-CAM heatmaps based on the average self-attention maps over the image patches classified to targeted textual tokens (i.e., the textual token(s) corresponding to the class label in the ImageNet dataset) in the last layer of the image encoder. We average the heatmaps over all attention heads.
As can be seen, our proposed model learns meaningful alignment between image patches and textual tokens.

\subsection{Prompt Templates for downstream tasks}
\label{apdx:prompt_template}
\paragraph{Image Classification.} Table~\ref{tbl:prompt_templates} shows the prompt templates for different image classification datasets
in the form of ``
$
\text{[prefix] \{label\}, [category description]. [suffix].} 
$
''
in Equation (\ref{eq:prompt_template}).
There are three components to be determined in the template, i.e., the prefix, the category description and the suffix.
For each component, we select several well-performed ones  for each dataset. 
Then we use the full combinations of all three components as the set of prompt templates for ensemble.
For instance, we use 5 prefixes, no category descriptions, and 6 suffixes for dataset ImageNet. 
Then the total number of prompt templates for this dataset is: $5 \times 1 \times 6=30$.

\paragraph{Image-text Retrieval.}  Following CLIP \citep{radford2021learning}, we use prompt in zero-shot image-text retrieval for both Flickr30K and MSCOCO datasets.
The prompt is selected by the same rule as described in Section~\ref{sec:prompt_ensemble}, except that we do not use ``[category description]'' here. Table \ref{tab:prompts_for_retrieval} shows the prompt templates for zero-shot image-text retrieval on Flickr30K and MSCOCO datasets.

\begin{table}
\small
    \centering
    \caption{Prompt templates used for 12 downstream image classification tasks. 
    }
    \label{tbl:prompt_templates}
    \resizebox{\textwidth}{!}{
    \begin{tabular}{m{1.5cm}|m{4cm}|m{2cm}|m{4cm}}
    	\toprule
    	Dataset      & Prefix                                                                                                                                           & Category description                                             & Suffix                                                                                                                                                \\ \midrule
    	CIFAR10      & ``a photo of a", ``a jpeg photo of a", ``a painting of a", ``itap of a", ``graffiti of a", ``a cartoon", ``a doodle"                                                      & None                                                             & None, ``It's common in daily life", ``It's cute", ``It's ugly", ``It's weird", ``Hope you like it"                                                     \\ \midrule
    	CIFAR100     & ``a jpeg photo of a", ``a painting  of a", ``a good photo  of a", ``a bad photo  of a", ``a photo  of a", ``itap of a", ``a rendering  of a"                                              & None                                                             & None, ``It's common in daily life", ``It's beautiful", ``It's ugly", ``I like it", ``I take it today"                                                  \\ \midrule
    
    	Caltech101   & ``a photo  of a", ``a cropped photo  of a", ``a good photo  of a", ``a bad photo  of a"                                                                                  & None                                                             & None, ``I like it", ``I hate it", ``It's ugly", ``It's cute"                                                                                           \\ \midrule
    	Stanford-Car & ``a photo  of a", ``a close-up photo  of a", ``a good photo  of a", ``a bad photo  of a"                                                                                 & ``a type of car", ``a type of automobile"                        & ``I like it", ``It belongs to my friend", ``It's brand new", ``It's popular recently", ``It's important to me", ``I take it today"                    \\ \midrule
    	Flowers102   & ``a photo of a (many) ", ``a rendering of a (many) ", ``itap of a (many) "                                                                                                                & ``a type of flower", ``a type of bloom"                          & ``It's beautiful", ``It's from my best friend", ``It gives out a sweet perfume/fragrance"                                                             \\ \midrule
    	ImageNet       & ``a photo of a", "a good photo of a", ``a bad photo of a", ``a close-up photo of a", ``itap of a"                                                                         & None                                                             & ``I like it", ``It's common in daily life", ``It's not common in daily life", ``It's ugly", ``It's cute", ``It's beautiful"                           \\ \midrule
    	Food101      & ``a photo of my", ``a close-up photo  of my", ``itap  of my"                                                                                                         & ``a type of food", ``a type of nourishment"                      & ``I made it today", ``I like it", ``I hate it", ``It's delicious", ``It's with nice flavour", ``It's with terrible flavour", ``It's popular recently" \\ \midrule
    	SUN397       & ``a photo  of a", ``a good photo  of a", ``a bad photo  of a", ``a bright photo  of a", a dark photo  of a", ``a black and white photo  of a", ``a nice scene  of a", ``a terrible scene  of a" & None                                                             & None, ``I like it", ``I hate it", ``It's beautiful", ``It's common in daily life", ``It's important to me"                                             \\ \midrule
    	DTD          & ``itap of a", ``a close-up photo of a"                                                                                                                     & ``texture", ``surface", ``material"                              & None, ``It's out of style", ``It's popular in old days", ``It's ugly", ``It's beautiful"                                                               \\ \midrule
    	Aircrafts    & ``a photo of the", ``a close-up photo of the", ``a good photo of the ", ``a pixelated photo of the"                                                                           & ``a type of plane", ``a type of aircraft", ``a type of airliner" & None,``I like it", ``It's important to me", ``I take it today", ``Hope you like it"                                                                    \\ \midrule
    	Oxford Pet   & ``a photo of my", ``a low resolution photo of my", ``a good photo of my"                                                                                           & ``a type of pet", ``a type of dog or cat"                        & None, ``It's cute", ``It's important to me", ``I like it", ``It's beautiful"                                                                           \\ \midrule
    	EuroSAT      & ``a photo  of a", ``a painting of a", ``a cropped photo of a", ``a good photo of a", ``a blurry photo of a"                                                                 & None, ``an example of aerial or satellite images"                 & None, ``I like it", ``It's taken from an aircraft or some flying object", ``It's collected by imaging satellites"                                      \\ \bottomrule
    \end{tabular}
    }
\end{table}

\begin{table}[!t]
\small
    \centering
    \caption{Prompt templates used for zero-shot image-text retrieval on Flickr30K and MSCOCO datasets. 
    }
    \label{tab:prompts_for_retrieval}
    \resizebox{0.8\textwidth}{!}{
    \begin{tabular}{m{1.5cm}|m{3cm}|m{3cm}|m{1.5cm}}
    	\toprule
    	Dataset  & Task & Prefix  & Suffix \\ \midrule
    	\multirow{2}{*}{ Flickr30K } & image-to-text retrieval      & ``a good photo of the'' &  ``I hate it.''        \\ 
    	 & text-to-image retrieval      & ``a good photo of'' &  None        \\ \midrule
    	\multirow{2}{*}{ MSCOCO } & image-to-text retrieval      & ``a good photo of'' &  ``It is ugly.''        \\ 
    	 & text-to-image retrieval      & None &  None       \\ \bottomrule

    \end{tabular}
    }
\end{table}

\end{document}